\documentclass{article}

\usepackage{PRIMEarxiv}

\usepackage{authblk}

\usepackage[utf8]{inputenc} % allow utf-8 input
\usepackage[T1]{fontenc}    % use 8-bit T1 fonts
\usepackage{hyperref}       % hyperlinks
\usepackage{url}            % simple URL typesetting
\usepackage{booktabs}       % professional-quality tables
\usepackage{amsfonts}       % blackboard math symbols
\usepackage{nicefrac}       % compact symbols for 1/2, etc.
\usepackage{microtype}      % microtypography
\usepackage{lipsum}
\usepackage{fancyhdr}       % header
\usepackage{graphicx}       % graphics
\graphicspath{{media/}}     % organize your images and other figures under media/ folder

\usepackage{longtable}
\usepackage{amsmath}
\usepackage{tikz}
\usepackage{forest}
\usepackage{xcolor}
\usepackage{longtable}
\definecolor{rootcolor}{RGB}{30,80,140}
\definecolor{col1}{RGB}{52,120,180}
\definecolor{col2}{RGB}{220,100,30}
\definecolor{col3}{RGB}{60,150,80}
\definecolor{col4}{RGB}{160,50,130}
\definecolor{col5}{RGB}{180,140,20}
\definecolor{col6}{RGB}{180,60,60}
\definecolor{leafcolor}{RGB}{245,245,245}

\title{Representation learning from OCT images
}

  \author[1]{Hedi Tabia}
  \author[1]{Désiré Sidibé}
  \author[2]{Nawres Khlifa}
  \author[3]{Ahmed Tabia}
  \author[4]{Ines Rahmany}
  \author[1]{Noura Aboudi}
  \author[1,2]{Zainab Haddad}
  \author[2]{Hajer Khachnaoui}
  \author[5]{Hsouna Zgolli}
  
  \affil[1]{IBISC Univ. Evry Université Paris-Saclay}
  \affil[2]{University of Tunis El Manar}
  \affil[3]{ESIEE Paris - Université Gustave Eiffel}
  \affil[4]{FST Sidi-Bouzid - University of Kairouan}
  \affil[5]{Department A - Hedi Raies of Ophthalmology Institut}

\begin{document}
\maketitle

\begin{abstract}
		Optical Coherence Tomography (OCT) has become one of the most used imaging modality in ophthalmology. It provides high-resolution, non-invasive visualization of retinal microarchitecture. The automated analysis of OCT images through representation learning has emerged as a central research frontier. This has mainly been  driven by the clinical need to process large acquisition volumes. The objective is to reduce the reliance on expert annotation, and improve diagnostic consistency across devices and populations. This survey provides a comprehensive and structured review of representation learning methods for retinal OCT image analysis. It covers the period from early deep learning approaches to the most recent developments in foundation models and vision-language systems. We organize the literature along a principled taxonomy of learning paradigms, encompassing supervised learning with CNN-based and transformer-based architectures, self-supervised and semi-supervised methods, generative approaches, as well as 3D volumetric modeling, multimodal representation learning, and large-scale pretrained foundation models. For each paradigm, we analyze the core methodological contributions, identify persistent limitations, and trace the connections between successive approaches. We further provide a structured overview of publicly available OCT datasets, discuss evaluation protocol considerations, and present a unified problem formulation that situates each learning paradigm within a common mathematical framework. Building on this analysis, we identify and discuss the most pressing open research directions emerging in the literature. This includes volumetric foundation model pretraining, uncertainty-aware representation learning, federated and privacy-preserving training, fairness and bias mitigation, concept-based interpretability, language-grounded clinical reasoning, and longitudinal disease progression modeling. Our study reveals that despite remarkable progress, fundamental challenges related to annotation scarcity, volumetric modeling, clinical interpretability, cross-device generalization, and prospective clinical validation remain unresolved. This survey aims to serve as a reference for researchers and clinicians working at the intersection of deep learning and ophthalmic imaging. It provides both a systematic account of the state of the art and a roadmap for the next generation of clinically deployable OCT analysis systems.
\end{abstract}

% keywords can be removed

\section{Introduction}

Optical Coherence Tomography~\cite{huang1991optical} (OCT) is a non-invasive, high-resolution imaging modality that has transformed the diagnosis and management of ophthalmic diseases. It exploits low-coherence interferometry to capture cross-sectional images of biological tissue at micrometer-scale resolution, enabling volumetric visualization of retinal layers, corneal structure, and other ocular microarchitectures that are imperceptible through conventional imaging. Since its clinical introduction in the early 1990s, OCT has become the standard diagnostic tool in ophthalmology~\cite{schuman2024optical}. Its applications span the detection of age-related macular degeneration (AMD)~\cite{hee1996optical}, diabetic macular edema (DME)~\cite{browning2004comparison}, glaucoma~\cite{jaffe2004optical}, retinal vein occlusion~\cite{rosenfeld2005optical}, and a growing spectrum of posterior and anterior segment pathologies~\cite{hwang2020accuracy,wong2026anterior}.

Despite its clinical success, OCT image analysis presents significant challenges. Modern acquisitions may consist of hundreds of cross-sectional slices (B-scans) per patient, placing substantial demands on clinician time and attention. OCT images are inherently affected by noise artifacts such as speckle, motion blur, and signal attenuation, which complicate the interpretation of fine structural details. Manual grading by expert clinicians remains the reference standard, yet it is time-consuming, expensive, and subject to inter- and intra-observer variability. Furthermore, it does not scale efficiently to large-scale screening programs. These limitations have motivated a rapidly growing body of research on automated computational methods for OCT image analysis.

OCT data exhibit intrinsic properties that make automated analysis particularly challenging. Speckle noise~\cite{schmitt1999speckle,szkulmowski2012efficient}, arising from coherent interference of backscattered light, degrades image quality and obscures subtle pathological features. Domain variability~\cite{vizzeri2009agreement,wu2023clinical,ndipenoch2024performance,ho2009assessment}, caused by differences in acquisition devices, protocols, and patient populations, limits model generalization. In addition, the scarcity of expert-annotated datasets, especially for volumetric data and rare pathologies, restricts the applicability of purely supervised approaches. The high dimensionality of 3D OCT volumes and the need to capture fine anatomical structures further complicate the learning process.

The emergence of deep learning and, more broadly, representation learning has fundamentally transformed OCT image analysis. Traditional approaches rely on handcrafted features engineered with domain expertise, whereas representation learning enables models to automatically discover hierarchical and semantically meaningful features directly from raw data. Convolutional neural networks~\cite{lecun2002gradient} (CNNs), vision transformers~\cite{dosovitskiy2020image} (ViTs), autoencoders~\cite{hinton2011transforming}, generative adversarial networks~\cite{goodfellow2020generative} (GANs), and self-supervised learning frameworks~\cite{liu2021self} have demonstrated remarkable performance across a wide range of OCT tasks, including segmentation, classification, lesion detection, denoising, super-resolution, and disease progression modeling.

Within this context, representation learning plays a central role in determining the performance, robustness, and clinical reliability of OCT-based systems. However, many existing approaches largely adopt methodologies originally developed for natural RGB images, relying on generic architectures and data-driven learning to implicitly capture OCT-specific characteristics. While effective in practice, this strategy often overlooks domain-specific properties such as retinal layer organization, volumetric coherence, and imaging physics, potentially limiting generalization and interpretability.

To address these challenges, the field has undergone a significant paradigm shift in recent years. The dominance of fully supervised learning has progressively given way to self-supervised approaches that leverage large amounts of unlabeled OCT data through contrastive learning and masked modeling. In parallel, generative models, including variational autoencoders and diffusion models, have emerged as powerful tools for learning structured latent representations and modeling data distributions. More recently, large-scale foundation models pretrained on multimodal datasets have demonstrated strong transferability and data efficiency, signaling a transition toward unified and task-agnostic representations. Additionally, there is a growing emphasis on exploiting 3D volumetric information and integrating complementary imaging modalities such as fundus photography, fluorescein angiography, and OCT angiography (OCTA) to better capture the complexity of retinal diseases.

Despite these advances, the field still lacks a comprehensive and structured synthesis of representation learning approaches for OCT imaging. Existing surveys often focus on specific tasks or architectures, without providing a unified perspective on the different learning paradigms and their interrelations. In this work, we address this gap by presenting a taxonomy-driven review of representation learning for OCT images. We systematically analyze supervised, self-supervised, generative, volumetric, multimodal, and foundation model approaches, highlighting their strengths, limitations, and underlying assumptions. We also review publicly available datasets and evaluation protocols, and identify key open challenges and promising future research directions. Our goal is to provide a coherent framework and roadmap for the development of robust, interpretable, and clinically reliable OCT analysis systems.

This survey is organized as follows. Section~\ref{sec:background} establishes the foundations of representation learning for OCT imaging. It introduces the formal problem setup and mathematical notation used throughout the review, identifies the key challenges specific to OCT data including noise, domain variability, and annotation scarcity, traces the historical transition from classical image processing to modern deep representation learning, surveys the publicly available OCT and OCTA datasets, discusses evaluation protocol considerations, and presents the hierarchical taxonomy of representation learning approaches that structures the remainder of the review. Sections~\ref{sec:supervised} through~\ref{sec:foundation} constitute the core of the survey, each devoted to a major learning paradigm. Section~\ref{sec:supervised} covers supervised representation learning, examining CNN-based architectures that learn discriminative local features from labeled B-scans, transformer-based architectures that exploit self-attention for global context modeling, and hybrid CNN-transformer designs that seek to combine the strengths of both. Section~\ref{sec:ssl} reviews self-supervised representation learning, covering contrastive, reconstruction-based, and pretext task approaches that leverage unlabeled OCT data to reduce annotation dependence. Section~\ref{sec:generative} addresses generative representation learning across three successive paradigms: autoencoders and variational models, GAN-based approaches, and the more recently emerged diffusion-based models, each offering distinct mechanisms for learning structured latent representations of retinal images. Section~\ref{sec:3d} examines 3D and volumetric representation learning, focusing on methods that exploit the full spatial continuity of OCT volumes rather than processing B-scans independently. Section~\ref{sec:multimodal} reviews multimodal representation learning, covering approaches that integrate OCT with complementary modalities such as fundus photography, OCTA, and clinical metadata to enrich learned representations. Section~\ref{sec:foundation} discusses foundation models and large-scale pretraining, examining vision-language models, contrastive multimodal pretraining, and domain-adapted ophthalmic foundation models that aim to learn universal representations transferable across tasks and modalities. Section~\ref{sec:challenges} synthesizes the persistent challenges and open research directions identified across all reviewed paradigms, organized into data limitations, representation learning deficiencies, and methodological gaps, and maps these challenges to the most promising future research directions including volumetric foundation model pretraining, uncertainty quantification, federated learning, fairness, concept-based interpretability, and longitudinal modeling. Finally, Section~\ref{sec:conclusion} concludes the survey with a synthesis of the main findings and a perspective on the trajectory of the field.

\begin{figure}[h]
	\includegraphics[width=\linewidth]{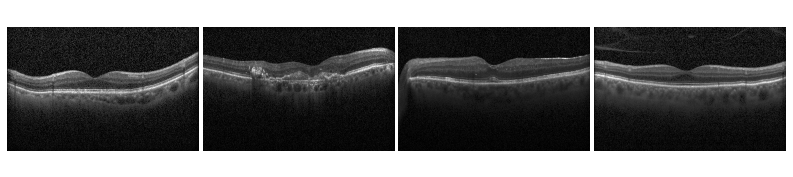}
	\caption{From left to right, four OCT B-scans corresponding to: drusen, choroidal neovascularization, diabetic macular edema, and normal. The images are from Kermany et al. \cite{kermany2018identifying} dataset.}
	\label{fig:sampleOctImages}
\end{figure}

\section{Background}
\label{sec:background}
OCT is based on low-coherence interferometry, conceptually analogous to ultrasound imaging but using light instead of sound. OCT measures the echo time delay and intensity of backscattered light to generate micrometer-resolution, cross-sectional images of biological tissues. In ophthalmology, OCT has become indispensable because it allows in vivo visualization of retinal layers, enabling early detection and monitoring of diseases such as Glaucoma, Age-related macular degeneration, and Diabetic retinopathy. Advances such as spectral-domain OCT (SD-OCT) and swept-source OCT (SS-OCT) have further improved imaging speed and depth. Figure \ref{fig:sampleOctImages} presents four samples of OCT images from Kermany et al. \cite{kermany2018identifying} dataset.

Representation learning for OCT images is the process of automatically learning meaningful, compact, and robust feature representations from raw OCT data that capture underlying anatomical structures, pathological patterns, and imaging characteristics. Instead of relying on handcrafted features, modern approaches use deep learning models to extract hierarchical representations that are optimized for downstream clinical tasks such as classification, segmentation, and diagnosis. These representations aim to be invariant to noise, acquisition variability, and patient-specific differences while preserving clinically relevant information.

\subsection{Problem Setup}
Let the OCT dataset be defined as:
\begin{equation}
	\mathcal{D} = \{(x_i, y_i)\}_{i=1}^N,
\end{equation}

where $x_i \in \mathbb{R}^{H \times W}$ (2D B-scan) or $x_i \in \mathbb{R}^{H \times W \times D}$ (3D OCT volume),
and $y_i \in \mathcal{Y}$ denotes the associated label (e.g., segmentation mask, class label, or clinical variable).

We assume that samples are drawn from an unknown data distribution:
\begin{equation}
	x \sim p_{\text{data}}(x).
\end{equation}

Due to the OCT acquisition process, observations are corrupted by noise (e.g., speckle noise), and can be modeled as:
\begin{equation}
	x = x^{\star} + \epsilon,
	\label{eq:noisemodel}
\end{equation}

where
$x^{\star}$ is the underlying clean anatomical structure, and $\epsilon$ which has the same dimension as $x$, is a stochastic noise term. Note that the speckle noise \cite{schmitt1999speckle} is a multiplicative noise pattern but often modeled using statistical distributions depending on the imaging conditions. A common transformation is to take the logarithm \cite{li2017statistical}, which converts the multiplicative model into an additive noise model.

The representation learning objective is to find a parametric mapping:
\begin{equation}
	f_{\theta} : \mathcal{X} \rightarrow \mathcal{Z}.
\end{equation}

Here $\mathcal{X}$ is the input space of OCT images, $\mathcal{Z} \subset \mathbb{R}^d$ is a latent representation space, and $\theta$ are the learnable parameters.

The learned representation is defined as:
\begin{equation}
	z = f_{\theta}(x).
\end{equation}

The objective is to learn $f_{\theta}$ such that $z$ captures relevant semantic information such as anatomical structures (e.g., retinal layers), pathological patterns, invariance to noise and acquisition variability.

In a general learning framework, the representation is optimized through a downstream task:
\begin{equation}
	\min_{\theta, \phi} \; \mathbb{E}_{(x,y)\sim \mathcal{D}} 
	\left[ \mathcal{L}\big(g_{\phi}(f_{\theta}(x)), y\big) \right],
	\label{eq:representationLearningSupervised}
\end{equation}

where $g_{\phi}$ is a task-specific predictor, and $\mathcal{L}(\cdot)$ is a task-dependent loss function.

More details are given in the following sections.

\subsection{Key Challenges in representation learning for OCT Images}
Representation learning for OCT images faces several unique challenges that distinguish it from standard RGB image analysis. First, OCT data are inherently affected by speckle noise \cite{schmitt1999speckle}, a multiplicative noise pattern that significantly degrades image quality and complicates feature extraction \cite{darlow2015review}.  This noise is often modeled using statistical distributions depending on the imaging conditions. A common transformation is to take the logarithm \cite{li2017statistical}, which converts the multiplicative model into an additive noise model, making it easier to handle in optimization and learning frameworks like the mostly additive noise in RGB images (Equation~\ref{eq:noisemodel}). Second, OCT images exhibit low contrast and subtle structural variations, requiring models to capture fine-grained anatomical details such as retinal layers, which are less prominent than objects in natural images. Third, there is a scarcity of annotated data, as labeling OCT requires expert clinicians, whereas large-scale labeled RGB datasets are widely available. Additionally, OCT data may also come as 3D volumetric scans, introducing higher computational complexity and the need to model spatial continuity across slices, unlike typical 2D RGB inputs. Another major challenge is domain shift, caused by differences in acquisition devices, protocols, and patient populations, which affects model generalization more severely than in RGB settings. Furthermore, OCT representations must be clinically interpretable, as decisions impact diagnosis, unlike many RGB tasks where interpretability is less critical. The presence of imaging artifacts such as shadowing and signal attenuation also complicates learning robust features. Moreover, OCT analysis often requires capturing geometric and layered structures, which are domain-specific and not commonly encountered in natural images. The need for robustness to small pathological changes and uncertainty estimation makes representation learning for OCT both more sensitive and more demanding than for standard images.

Interestingly, despite the specific challenges associated with OCT imaging, the majority of representation learning methods developed in recent years largely follow the same paradigms as those used for standard RGB images. In most cases, models are directly adapted from computer vision, relying on architectures such as CNNs or Transformers originally designed for natural images, with minimal structural modifications. The underlying assumption is that these models can implicitly learn OCT-specific characteristics such as speckle noise, layered anatomical structures, and low-contrast patterns purely from the data. As a result, researchers often depend heavily on data-driven learning, using augmentations or large-scale pretraining to encourage robustness, rather than explicitly integrating domain knowledge about OCT physics or retinal geometry. While this strategy has led to strong empirical performance, it also reveals a limitation: the representations are not inherently tailored to the unique properties of OCT data but instead attempt to approximate them through general-purpose feature extractors.

\subsection{From classical image processing to deep representation learning}

Classical methods for OCT image analysis \cite{abramoff2010retinal,baghaie2015state,alsaih2017machine} rely on traditional image processing techniques such as median or wavelet filtering for denoising, edge detection and gradient-based approaches for layer segmentation, graph-based strategies like shortest path or dynamic programming, and model-driven formulations including active contours and level sets. These approaches are generally interpretable, mathematically well-grounded, and require relatively limited data, making them effective in controlled environments. However, these methods remain highly sensitive to speckle noise and imaging artifacts. They depend on careful parameter tuning. Their generalization across datasets and devices is limited. Performance also degrades in the presence of irregular or atypical pathological structures. In contrast, the advent of deep neural networks (DNNs) has fundamentally transformed OCT analysis. Convolutional neural networks in particular enable automatic learning of hierarchical representations directly from raw imaging data. Unlike hand-crafted approaches, these models do not require manual feature engineering. They discover relevant features autonomously through end-to-end training on labeled datasets. These models achieve state-of-the-art performance in tasks such as segmentation, classification, and anomaly detection, while offering robustness to anatomical and acquisition variability when trained on sufficiently diverse datasets. They also support end-to-end learning pipelines and can leverage 3D volumetric context through advanced architectures such as 3D CNNs and transformers. Nevertheless, DNNs introduce their own limitations, including a strong dependence on large annotated datasets, limited interpretability often perceived as a “black box” issue in clinical contexts, susceptibility to domain shift when applied to unseen data, and significant computational cost, particularly for volumetric processing. Moreover, their integration into clinical practice requires rigorous validation, explainability, and compliance with regulatory standards, highlighting an ongoing challenge in bridging performance and trust.

\subsection{Public OCT Datasets}

The development of representation learning methods for OCT images has been made possible by the progressive release of public datasets, summarized in Table~\ref{tab:OCTdatasets}. However, unlike standard image analysis, where large-scale unified benchmarks such as ImageNet have structured the field around a common evaluation framework, OCT research relies on a heterogeneous collection of datasets that differ substantially in scale, modality, annotation type, acquisition device, and targeted pathology. Classification datasets constitute the most widely used category. The Kermany et al.\cite{kermany2018identifying} dataset remains the de facto standard for classification benchmarking, providing over 207,000 labeled B-scans across four categories acquired with a single Spectralis device, and its scale has driven many of the early deep learning advances in retinal disease recognition. More recent classification datasets such as OCTDL \cite{kulyabin2024octdl} and OCT-C8 \cite{subramanian2022classification} extend the disease spectrum beyond the canonical four categories of the Kermany benchmark, covering conditions including epiretinal membrane, retinal vein and artery occlusions, and vitreomacular interface disease, thereby enabling evaluation on a broader and more clinically representative range of pathologies. Segmentation datasets address a complementary and more annotation-intensive challenge. The Duke DME \cite{chiu2015kernel} dataset, though limited to 110 B-scans from 10 subjects, established an early standard for retinal layer boundary annotation. RETOUCH \cite{bogunovic2019retouch} subsequently introduced multi-vendor variability by providing fluid segmentation labels across three different OCT devices, making it a critical resource for studying domain generalization in segmentation models. AROI \cite{melinscak2021aroi} extended this direction by providing joint layer and fluid annotations in neovascular AMD, while OIMHS \cite{ye2023oimhs} contributed a large single-disease segmentation dataset focused on macular holes with four distinct label types. The recently released OCT5k \cite{arikan2025oct5k} dataset represents the most comprehensive segmentation resource to date, combining multi-grader pixel-wise layer annotations with biomarker bounding box labels across three disease categories, and stands as the first dataset to systematically address inter-grader variability at scale. Volumetric modeling is partially supported by datasets providing 3D OCT volumes, including RETOUCH \cite{bogunovic2019retouch} and OCT5k \cite{arikan2025oct5k}, yet fully annotated 3D datasets remain scarce relative to their 2D counterparts, reflecting the substantially higher cost of volumetric labeling and constraining the development of architectures designed to exploit inter-slice continuity. Multimodal datasets represent a more recent and clinically motivated development. GAMMA \cite{wu2023gamma} provides paired fundus and OCT volumes for glaucoma, while MultiEYE \cite{wang2024multieye} constitutes the largest publicly available paired multimodal dataset, combining over 58,000 fundus photographs with nearly 46,000 OCT B-scans across nine disease categories. These resources are essential for the multimodal and foundation model approaches reviewed in later sections, as they enable the study of cross-modal representation learning and joint feature fusion. The OCTA-500 \cite{li2024octa} dataset extends the landscape further by providing the largest public OCTA resource, with over 360,000 scans from 500 subjects under two fields of view, and rich vascular annotations covering artery, vein, capillary, and foveal avascular zone segmentation.

Despite this collective progress, several structural limitations persist across the available dataset landscape. Acquisition heterogeneity is pervasive: datasets are collected from different devices Spectralis, Cirrus, Triton, Optovue each with distinct resolution, scanning protocol, and signal characteristics, introducing domain shifts that challenge model generalization and make cross-dataset comparison unreliable. Annotation heterogeneity compounds this problem, as label types range from image-level class labels to dense pixel-wise segmentation masks, with inter-expert variability affecting ground truth reliability particularly in segmentation tasks where layer boundary placement is inherently ambiguous. The predominance of 2D B-scan datasets over fully annotated 3D volumes creates an inconsistency between the data on which models are trained and the volumetric nature of OCT acquisitions, limiting the validity of benchmark evaluations. Disease prevalence imbalance is another recurring concern: most datasets are dominated by AMD and DME, with rare pathologies severely underrepresented, biasing both training and evaluation toward common conditions. Finally, dataset scale remains a limiting factor for data-hungry paradigms including Vision Transformers, diffusion models, and foundation models, as the cost of expert annotation constrains most datasets to sizes orders of magnitude smaller than their natural image counterparts. Taken together, these characteristics underscore that the absence of a unified, large-scale, multi-device, and multi-task OCT benchmark represents one of the most significant infrastructure gaps in the field, and that progress on standardization of evaluation protocols is as necessary as advances in model architecture.

\subsection{Evaluation Protocol Considerations}

Ensuring fair and reproducible evaluation in OCT-based representation learning requires careful attention to several methodological considerations that are frequently overlooked in the literature. The most critical is patient-level data splitting: scans from the same patient must not appear in both training and test sets, as random B-scan-level splitting allows the model to memorize patient-specific retinal features rather than learning generalizable representations, leading to artificially inflated performance estimates that do not reflect real-world diagnostic capability. This issue is particularly acute in datasets such as Kermany et al. \cite{kermany2018identifying}, where the large number of B-scans per patient makes random splitting especially prone to leakage. Beyond within-dataset evaluation, cross dataset evaluation is essential for measuring generalization across acquisition devices, scanning protocols, and patient populations: a model trained on Spectralis data and evaluated only on Spectralis data provides no evidence of robustness to the domain shifts that will inevitably be encountered in clinical deployment, and evaluation across datasets such as RETOUCH, which spans three different vendors, provides a more clinically meaningful measure of performance. The metrics selection must be adapted to the task at hand. For classification, accuracy alone is insufficient given the class imbalance characteristic of most OCT datasets; area under the ROC curve (AUC), F1-score, and per-class sensitivity and specificity provide a more complete picture of model behavior across disease categories. For segmentation, the Dice coefficient and Intersection over Union (IoU) measure region overlap, while the Hausdorff distance captures boundary localization accuracy, which is particularly relevant for retinal layer delineation where clinically meaningful measurements depend on precise boundary placement rather than bulk region overlap. For volumetric datasets, evaluation must additionally account for 3D spatial coherence: metrics computed independently on individual B-scans do not penalize segmentations that are locally accurate but globally inconsistent across slices, and volume-level aggregation or 3D variants of standard metrics are therefore preferable. Finally, domain adaptation and robustness protocols deserve systematic integration into standard evaluation pipelines. Models should be assessed under controlled distribution shifts ; for instance, training on one device and testing on another, or training on one demographic cohort and testing on another. This is important to quantify the degree to which learned representations are device-invariant and population-generalizable. The absence of these protocols from the majority of reviewed works is itself a significant methodological gap, and their adoption as standard practice would substantially improve the comparability and clinical relevance of reported results across the field.

%	\begin{table*}[htbp]
	%		\centering
	%		\caption{Summary of publicly available retinal OCT and OCTA datasets.
		%			\textbf{Dim.}: dimensionality (2D B-scans or 3D volumes);
		%			\textbf{Task}: Cls = Classification, Seg = Segmentation, Det = Detection, Gen = Generation/Augmentation;
		%			\textbf{Annotations}: CL = Class Labels, RL = Retinal Layers, FL = Fluid Labels, BBox = Bounding Boxes, Seg = Pixel-wise Segmentation masks.
		%			AMD = Age-related Macular Degeneration, DME = Diabetic Macular Edema, CNV = Choroidal Neovascularization,
		%			CSR = Central Serous Retinopathy, DR = Diabetic Retinopathy, MH = Macular Hole, MS = Multiple Sclerosis,
		%			RVO = Retinal Vein Occlusion, RAO = Retinal Artery Occlusion, ERM = Epiretinal Membrane,
		%			VID = Vitreomacular Interface Disease, PED = Pigment Epithelial Detachment.}
	%		\resizebox{\textwidth}{!}{%
		%			\begin{tabular}{p{3cm}p{2cm}lp{1cm}p{2cm}p{3cm}p{7.5cm}}
			%	\begin{adjustbox}{width=\textwidth}
				{\small
					\begin{longtable}{p{1.5cm}p{1.5cm}p{1.3cm}p{1.3cm}p{2.5cm}p{2cm}p{5cm}}
						\caption{Summary of publicly available retinal OCT and OCTA datasets.
							\textbf{Dim.}: dimensionality (2D B-scans or 3D volumes);
							\textbf{Task}: Cls = Classification, Seg = Segmentation, Det = Detection, Gen = Generation/Augmentation;
							\textbf{Annotations}: CL = Class Labels, RL = Retinal Layers, FL = Fluid Labels, BBox = Bounding Boxes, Seg = Pixel-wise Segmentation masks.
							AMD = Age-related Macular Degeneration, DME = Diabetic Macular Edema, CNV = Choroidal Neovascularization,
							CSR = Central Serous Retinopathy, DR = Diabetic Retinopathy, MH = Macular Hole, MS = Multiple Sclerosis,
							RVO = Retinal Vein Occlusion, RAO = Retinal Artery Occlusion, ERM = Epiretinal Membrane,
							VID = Vitreomacular Interface Disease, PED = Pigment Epithelial Detachment.}\\
						\toprule
						\textbf{Dataset} & \textbf{Size} & \textbf{Dim.} & \textbf{Task} & \textbf{Annotations} & \textbf{Diseases} & \textbf{Remarks} \\
						\midrule
						\midrule
						
						Kermany et al. \cite{kermany2018identifying}
						& 207,130 B-scans
						& 2D
						& Cls
						& CL
						& CNV, DME, Drusen, Normal
						& Largest public OCT dataset. Tiered expert grading. Single-device (Spectralis). Widely used benchmark. Available on Mendeley Data. \\
						\midrule
						\addlinespace
						
						Srinivasan et al. \cite{srinivasan2014fully}
						& 45 volumes (3 categories)
						& 2D/3D
						& Cls
						& CL
						& AMD, DME, Normal
						& Early benchmark dataset. Small scale. Used for automated disease detection. \\
						\midrule
						\addlinespace
						OCTID \cite{gholami2020octid}
						& $>$500 B-scans
						& 2D
						& Cls, Seg
						& CL, RL (25 images)
						& AMD, MH, CSR, DR, Normal
						& Acquired with Cirrus HD-OCT. Includes 25 manually segmented normal images. GUI provided for semi-automated segmentation. \\
						\midrule
						\addlinespace
						
						OCTDL \cite{kulyabin2024octdl}
						& 2,064 B-scans
						& 2D
						& Cls
						& CL
						& AMD, DME, ERM, RAO, RVO, VID, Normal
						& Broader disease spectrum than Kermany. Acquired with Optovue Avanti RTVue XR. Second largest public OCT classification dataset. \\
						\midrule
						\addlinespace
						
						OCT-C8 \cite{subramanian2022classification}
						& 24,000 B-scans
						& 2D
						& Cls
						& CL
						& AMD, CNV, CSR, DME, MH, Drusen, DR, Normal
						& Eight-class dataset. Pre-divided into train/val/test splits. \\
						\midrule
						\addlinespace
						
						OCT MS and HC \cite{he2019retinal}
						& 35 subjects
						& 2D/3D
						& Cls, Seg
						& CL, RL
						& MS, Normal
						& Provided by Johns Hopkins University. Includes healthy controls and MS patients. Requires preprocessing for segmentation evaluation. \\
						
						\midrule
						Duke DME \cite{chiu2015kernel}
						& 110 B-scans (10 subjects)
						& 2D
						& Seg
						& RL (8 boundaries)
						& DME
						& Acquired with Heidelberg Spectralis. Expert-annotated layer boundaries. Widely used for segmentation benchmarking. \\
						\midrule
						\addlinespace
						
						RETOUCH \cite{bogunovic2019retouch}
						& 70 volumes
						& 3D
						& Seg
						& FL (IRF, SRF, PED)
						& AMD, RVO
						& Multi-vendor dataset (Cirrus, Triton, Spectralis). Annotated at Medical University of Vienna and Radboud UMC. Challenge benchmark. \\
						\midrule
						\addlinespace
						
						UMN (University of Minnesota) \cite{rashno2017fully}
						& 600 B-scans
						& 2D
						& Seg
						& FL (IRF, SRF, PED)
						& AMD (exudative)
						& $\sim$100 B-scans per subject. Manual fluid annotations. Used for fluid segmentation validation. \\
						\midrule
						\addlinespace
						
						AROI \cite{melinscak2021aroi}
						& 1,136 B-scans (24 subjects)
						& 2D
						& Seg
						& RL, FL
						& AMD (nAMD)
						& Acquired with Zeiss Cirrus HD OCT 4000. Expert pixel-wise annotations for layers and fluid. Inter-observer agreement reported. \\
						\midrule
						\addlinespace
						
						OIMHS \cite{ye2023oimhs}
						& 3,859 B-scans (119 subjects)
						& 2D
						& Seg
						& Seg (4 labels)
						& MH
						& Four segmentation labels: retina, macular hole, intraretinal cysts, choroid. Large single-disease dataset. \\
						\midrule
						\addlinespace
						
						OCT5k \cite{arikan2025oct5k}
						& 1,672 scans (5,016 labels)
						& 2D/3D
						& Seg, Det
						& RL (5 boundaries), BBox (9 classes)
						& AMD, DME, Normal
						& Multi-grader annotations. Largest multi-disease layer segmentation dataset. Additional biomarker bounding box labels. \\
						
						\midrule
						FUND-OCT \cite{hassan2022composite}
						& 105 subjects
						& 2D
						& Cls
						& CL
						& 4 disease types
						& Paired fundus and OCT images. Small scale. Limited disease diversity. \\
						\midrule
						\addlinespace
						
						MMC-AMD \cite{wang2022learning}
						& $<$2,000 images
						& 2D
						& Cls
						& CL
						& AMD
						& Paired multimodal images. Single disease. Limited scale. \\
						\midrule
						\addlinespace
						
						GAMMA \cite{wu2023gamma}
						& 300 samples
						& 2D/3D
						& Cls, Seg
						& CL, RL
						& Glaucoma
						& Paired fundus photographs and OCT volumes. Used in glaucoma detection challenges. \\
						\midrule
						\addlinespace
						
						MultiEYE \cite{wang2024multieye}
						& 58,036 fundus + 45,923 OCT B-scans
						& 2D
						& Cls
						& CL
						& AMD, DR, Glaucoma, Myopia, MEM, CSC, and others (9 classes)
						& Largest public multimodal OCT/fundus dataset. Patient-wise splits. Multi-class. Unpaired multimodal design. \\
						
						\midrule
						OCTA-500 \cite{li2024octa}
						& 500 subjects (361,600 scans)
						& 2D/3D
						& Seg, Cls
						& RL, vessel (7 label types), CL
						& DR, AMD, Glaucoma, CNV, Normal
						& Largest public OCTA dataset ($>$80 GB). Two fields of view (3mm/6mm). OCT and OCTA volumes. Rich annotations including artery/vein/capillary/FAZ segmentation. \\
						
						\midrule
						RASTA \cite{germanese2023retinal}
						& 499 patients 814 volumes 2005 en face
						& 3D/2D
						& Cls, Pred
						& CL
						& Cardiovascular risk
						& Multimodal dataset (SS-OCTA volumes + en face + clinical data). Designed for systemic risk prediction (oculomics). Moderate scale. Single-center acquisition. Weak labels based on clinical risk scores (CHA$_2$DS$_2$-VASc). \\
						
						\midrule
						ROSE \cite{ma2020rose}
						& 229 images (117 ROSE-1)
						& 2D
						& Seg
						& Seg (pixel + centerline)
						& Microvascular structure
						& First public OCTA vessel segmentation dataset. En face angiograms. Fine vessel annotations (including centerlines). Very small-scale. Single-task dataset. No volumetric information. \\
						
						\midrule
						SYN-OCT \cite{wong2026syn}
						& 200,000 Synthetic
						& 2D
						& Cls, Seg, Gen
						& RL, Seg
						& Glaucoma, Healthy
						& Synthetic OCT dataset with perfect annotations. Enables large-scale training and pretraining. Useful for data augmentation. Domain gap with real data limits direct clinical applicability. 
						\\
						
						\bottomrule
						%
						%	}
						\label{tab:OCTdatasets}
					%\end{table*}
			\end{longtable}}
			%	\end{adjustbox}

		\subsection{Taxonomy of Representation Learning for OCT Images}
		
\begin{figure*}[htbp]
			\centering
			\resizebox{0.9\textwidth}{!}{%
				\begin{forest}
					for tree={
						grow=east,
						reversed=true,
						anchor=base west,
						parent anchor=east,
						child anchor=west,
						base=left,
						font=\small\sffamily,
						rectangle,
						draw=gray!60,
						rounded corners=2pt,
						align=left,
						inner sep=4pt,
						l sep=8mm,
						s sep=2mm,
						edge={gray, -latex},
					},
					where level=0{
						fill=rootcolor,
						text=white,
						font=\bfseries\sffamily,
						minimum width=3.2cm,
						minimum height=0.9cm,
						align=center,
					}{},
					where level=1{
						fill=col1,
						text=white,
						font=\bfseries\small\sffamily,
						minimum width=3.6cm,
					}{},
					where level=2{
						fill=col1!25,
						text=black,
						font=\bfseries\footnotesize\sffamily,
						minimum width=3.4cm,
					}{},
					where level=3{
						fill=leafcolor,
						text=black,
						font=\footnotesize\sffamily,
						draw=gray!40,
						minimum width=4.5cm,
					}{},
					[{Representation\\Learning\\for OCT}, align=center
					[{Learning\\Paradigm}, fill=col1, text=white, align=center
					[{Supervised}, fill=col1!30, align=left
					[{CNN-based}, fill=leafcolor]
					[{Encoder-Decoder (U-Net variants)}, fill=leafcolor]
					[{Transformer-based (ViT\, Swin\, Hybrids)}, fill=leafcolor]
					]
					[{Self-Supervised (SSL)}, fill=col1!30, align=left
					[{Contrastive (SimCLR\, MoCo\, BYOL)}, fill=leafcolor]
					[{Reconstruction (AE\, VAE\, MAE)}, fill=leafcolor]
					[{Pretext tasks (rotation\, inpainting)}, fill=leafcolor]
					]
					[{Generative}, fill=col1!30, align=left
					[{Variational Autoencoders (VAE)}, fill=leafcolor]
					[{Adversarial Networks (GAN)}, fill=leafcolor]
					[{Diffusion Models}, fill=leafcolor]
					]
					[{Foundation Models}, fill=col1!30, align=left
					[{Transfer from natural images}, fill=leafcolor]
					[{Medical foundation models}, fill=leafcolor]
					[{Domain-adapted OCT pretraining}, fill=leafcolor]
					]
					]
					[{Model\\Architecture}, fill=col2, text=white, align=center
					[{CNN-based}, fill=col2!30, align=left
					[{Local texture and spatial features}, fill=leafcolor]
					[{ResNet\, DenseNet\, VGG}, fill=leafcolor]
					]
					[{Transformer-based}, fill=col2!30, align=left
					[{Vision Transformer (ViT)\, Swin}, fill=leafcolor]
					[{CNN-Transformer hybrids}, fill=leafcolor]
					]
					[{3D Volumetric}, fill=col2!30, align=left
					[{3D CNNs\, 3D Transformers}, fill=leafcolor]
					[{Slice aggregation networks}, fill=leafcolor]
					]
					]
					[{Data\\Modality}, fill=col3, text=white, align=center
					[{Single-Modality OCT}, fill=col3!30, align=left
					[{2D B-scans}, fill=leafcolor]
					[{3D OCT volumes}, fill=leafcolor]
					]
					[{Multimodal}, fill=col3!30, align=left
					[{OCT + Fundus imaging}, fill=leafcolor]
					[{OCT + OCTA}, fill=leafcolor]
					[{OCT + Clinical metadata}, fill=leafcolor]
					]
					]
					[{Downstream\\Applications}, fill=col4, text=white, align=center
					[{Classification}, fill=col4!30, align=left
					[{AMD\, DME\, CNV detection}, fill=leafcolor]
					[{Multi-class disease diagnosis}, fill=leafcolor]
					]
					[{Segmentation}, fill=col4!30, align=left
					[{Retinal layer delineation}, fill=leafcolor]
					[{Fluid region segmentation}, fill=leafcolor]
					]
					[{Restoration}, fill=col4!30, align=left
					[{Denoising\, super-resolution}, fill=leafcolor]
					[{Artifact removal}, fill=leafcolor]
					]
					[{Prognosis}, fill=col4!30, align=left
					[{Disease progression modeling}, fill=leafcolor]
					[{Longitudinal analysis}, fill=leafcolor]
					]
					]
					]
				\end{forest}
			}
			\caption{Hierarchical taxonomy of representation learning methods for Optical Coherence Tomography (OCT) image analysis, organized along four complementary dimensions: learning paradigm, model architecture, data modality, and downstream application.}
			\label{fig:taxonomy}
		\end{figure*}
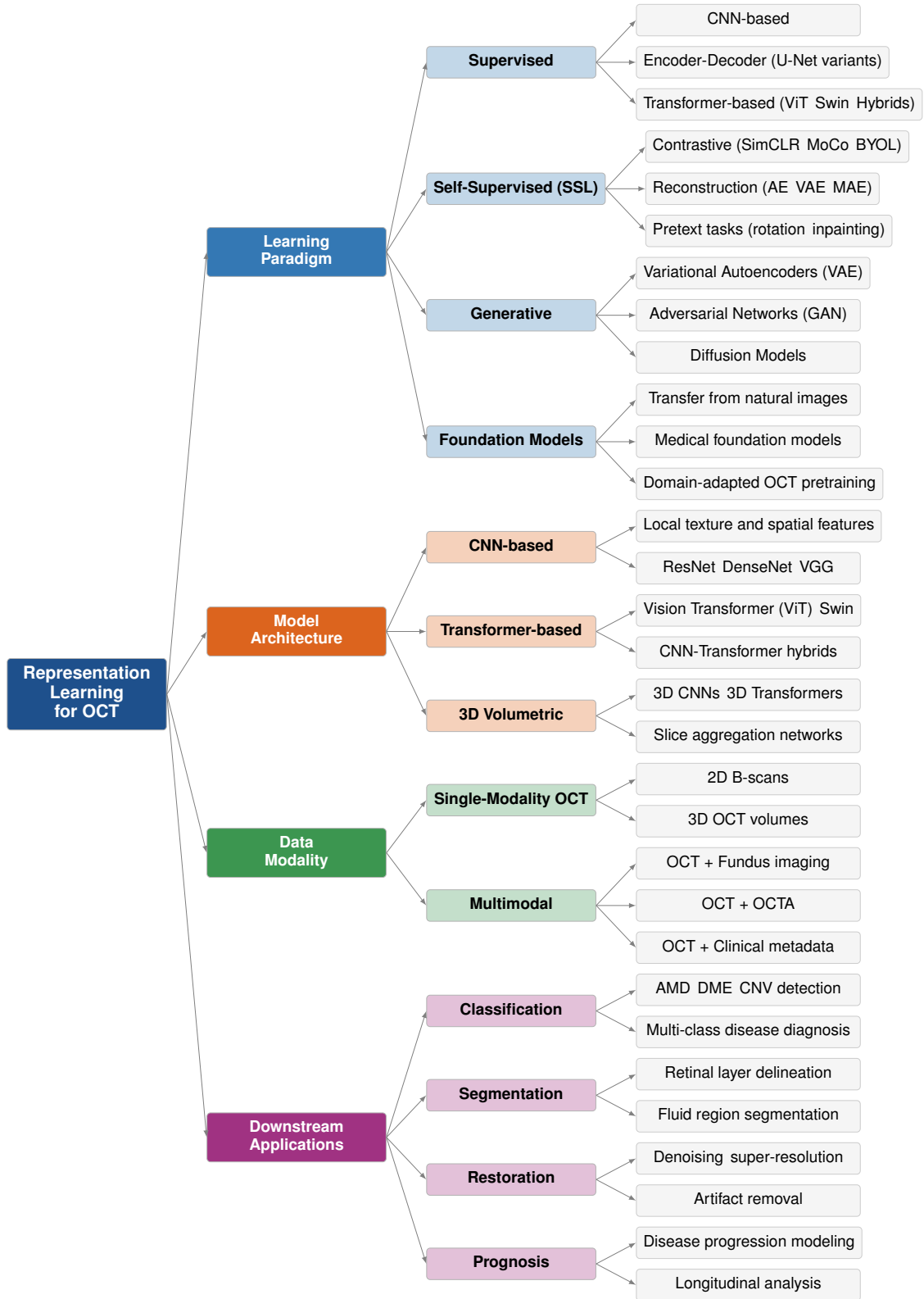

		Representation learning methods for OCT images can be organized along a hierarchical taxonomy that captures the diversity of approaches proposed in the literature, as illustrated in Figure~\ref{fig:taxonomy}. The first dimension concerns the \textit{learning paradigm}, which defines how the supervisory signal is obtained and how the representation is optimized. Supervised learning approaches train models end-to-end using manually annotated labels, relying on CNN-based architectures such as ResNet \cite{he2016deep} and DenseNet \cite{huang2017densely} or encoder-decoder designs such as U-Net \cite{ronneberger2015u} variants to learn discriminative features directly from labeled B-scans. Self-supervised learning methods instead construct supervisory signals from the data itself, through contrastive objectives such as SimCLR \cite{chen2020simple}, MoCo \cite{he2020momentum}, and BYOL \cite{grill2020bootstrap}, reconstruction-based objectives such as autoencoders, variational autoencoders, and masked image modeling, or pretext tasks such as rotation prediction, patch ordering, and inpainting. Generative approaches model the underlying data distribution explicitly, using variational autoencoders, generative adversarial networks, or the more recently introduced diffusion models, to learn structured latent representations that support synthesis, enhancement, and anomaly detection. Foundation and pretrained model approaches leverage large-scale pretraining, either from natural image datasets or from domain-specific medical corpora, and adapt the resulting representations to OCT tasks through transfer learning or domain-specific fine-tuning. The second dimension concerns \textit{model architecture}, which determines how spatial information is processed and at what scale features are extracted. CNN-based models capture local texture and spatial patterns through hierarchical convolution, transformer-based models including Vision Transformers and Swin Transformers capture long-range dependencies through self-attention, 3D volumetric architectures extend either paradigm to process full OCT volumes and exploit inter-slice continuity, and hybrid architectures combine convolutional and attention mechanisms to leverage the strengths of both. The third dimension concerns \textit{data modality}, reflecting the richness of information available during learning. Methods may operate on single-modality OCT data either individual 2D B-scans or full 3D volumes or integrate complementary modalities such as fundus photography, OCT angiography, and clinical metadata within multimodal or cross-modal representation learning frameworks. The fourth dimension concerns \textit{downstream application}, which determines the practical clinical objective that the learned representation must support. These applications include disease classification such as AMD, DME, and CNV detection; anatomical and pathological segmentation of retinal layers and fluid regions; image restoration tasks including denoising, super-resolution, and artifact removal; anomaly detection for rare or unseen pathologies; and longitudinal prognosis modeling for disease progression analysis. Although this taxonomy provides a comprehensive map of the field along all four dimensions simultaneously, the present review is primarily organized along the \textit{learning paradigm} axis. This choice reflects the most natural and informative structuring principle for the literature: the learning paradigm determines fundamental assumptions about data availability, annotation requirements, and the nature of the learned representation, and the transition across paradigms  from fully supervised to self-supervised, generative, and foundation model approaches traces the central methodological trajectory of the field over the past decade. Architectural, modality, and application considerations are discussed within each paradigm section as they arise, allowing the review to maintain a coherent narrative thread while preserving the cross-dimensional richness captured by the full taxonomy.

		\section{Supervised Representation Learning}
		\label{sec:supervised}
		In the fully supervised setting, the complete dataset $\mathcal{D} = \{(x_i, y_i)\}_{i=1}^N$ is assumed to be labeled, where each input $x_i \in \mathbb{R}^{H \times W}$ is a 2D OCT B-scan and $y_i \in \mathcal{Y}$ denotes its associated annotation. $y_i$ is  a discrete class label for classification or a dense segmentation mask for layer and lesion delineation. The goal is to learn a parametric encoder $f_{\theta} : \mathcal{X} \rightarrow \mathcal{Z}$ composed with a task-specific predictor $g_{\phi} : \mathcal{Z} \rightarrow \mathcal{Y}$, trained end-to-end by minimizing the empirical risk in Equation~\ref{eq:representationLearningSupervised}.
		% \begin{equation}
			% \min_{\theta, \phi} \; \mathbb{E}_{(x,y)\sim \mathcal{D}} \left[ \mathcal{L}\big(g_{\phi}(f_{\theta}(x)), y\big) \right]
			% \end{equation}
		In that equation the loss function $\mathcal{L}$ could be a cross-entropy for classification or Dice loss for segmentation. 
		
		CNN-based architectures instantiate $f_\theta$ as a hierarchy of convolutional layers that progressively extract local spatial features, while transformer-based and hybrid architectures augment or replace convolutional operations with self-attention mechanisms to additionally capture global dependencies. In all cases, the full annotation set is required at training time, making performance directly contingent on the quantity and quality of labeled data.
		
		\subsection{CNN-based Architectures}
		
		Convolutional Neural Networks \cite{lecun2002gradient} (CNNs) represent the earliest and most established paradigm for OCT-based retinal analysis. They learn discriminative hierarchical features directly from raw B-scans in a fully supervised manner. Works such as OctNet \cite{sunija2021octnet}, AOCT-Net \cite{alqudah2020aoct}, and other CNN-based classification frameworks \cite{awais2017classification,hosni2023prediction,rong2018surrogate,elkholy2024deep,haddad2025segmentation,mlaouhi2025advanced} adopt an end-to-end paradigm mapping OCT images directly to disease labels. These works demonstrate that CNNs effectively capture spatial features across multiple diagnostic categories. A complementary strategy is explored in \cite{raja2020extraction}, where deep learning is combined with anatomical domain knowledge: retinal layers are first segmented, and clinically relevant biomarkers such as the cup-to-disc ratio are derived for downstream disease detection.
		
		Despite strong empirical performance, CNN-based approaches exhibit several important limitations. Most methods are trained on datasets covering only a small number of disease categories typically choroidal neovascularization, diabetic macular edema, drusen, and normal cases. This narrow scope raises concerns about generalization to rarer pathologies and to data from different devices or clinical settings. Evaluation methodology is frequently suboptimal: random rather than patient-wise data splits can inflate performance estimates through data leakage. A more fundamental architectural limitation is the reliance on 2D CNNs operating on individual B-scans. While computationally efficient, this approach discards inter-slice correlations that carry critical diagnostic information. Patch-based segmentation variants further suffer from restricted receptive fields and limited global contextual awareness, negatively affecting layer boundary detection accuracy.
		Interpretability is another persistent concern. Post-hoc visualization techniques such as Grad-CAM \cite{selvaraju2017grad} and saliency maps \cite{etmann2019connection} have been integrated into CNN classifiers to provide some degree of decision transparency \cite{haddad2024explainable}. However, these methods do not alter the underlying representation learning strategy. The decision-making process of CNNs remains largely opaque, which is a significant barrier to clinical adoption where transparency and accountability are essential requirements. Finally, the predominant focus on OCT images alone, without integration of complementary sources such as fundus photography or patient metadata, limits the diagnostic richness of learned representations. Together, these limitations motivate the development of architectures incorporating volumetric modeling, global contextual reasoning, and multimodal information directions explored throughout the remainder of this review.
		
		\subsection{Transformer-based Architectures}
		
		A parallel line of supervised work has explored discriminative architectures based on the Transformer paradigm \cite{vaswani2017attention}, in particular Vision Transformers \cite{dosovitskiy2020image}. The core motivation is that CNNs are inherently limited in modeling long-range spatial dependencies due to their local receptive fields. Transformers address this through self-attention, which captures global contextual information across entire images. This offers a structural advantage for retinal analysis, where pathological features may be distributed across spatially distant regions. Several architectures have been specifically designed to adapt this paradigm to OCT data. SViT \cite{zhao2023svit,hemalakshmi2024automated}, OCTformer \cite{wang2023octformer}, CRAT \cite{yang2025crat}, and Oct-Trans \cite{elsharkawy2025oct} all employ hierarchical structures aimed at capturing complex pathological patterns at multiple scales.
		In classification, works including \cite{hemalakshmi2024automated,yang2025crat,cai2023classification} demonstrate that transformer-based models achieve competitive or superior performance compared to CNNs. \cite{kihara2022detection} further highlights their capacity to detect subtle features distributed across different regions of OCT scans. In segmentation, \cite{philippi2023vision} applies transformer-based encoders to improve lesion delineation through better pixel-level contextual modeling. Interpretability is partially addressed in \cite{he2023interpretable} through attention map visualization, offering a degree of decision transparency absent from CNN-based approaches. Beyond static analysis, MBT \cite{ait2023mbt} handles video-like OCT sequences to capture temporal dependencies. Multimodal extensions such as \cite{elsharkawy2025oct} further exploit transformers to fuse heterogeneous feature sources, demonstrating their flexibility across input types.
		Despite these advantages, transformer-based models carry limitations that are in several respects more acute than those of CNNs. They are highly data-hungry, requiring large-scale annotated datasets for optimal performance. This is particularly problematic in medical imaging, and directly motivates the self-supervised and generative approaches reviewed in subsequent sections. Computational cost is also substantially higher than CNNs. Self-attention scales quadratically with the number of input tokens, hindering application to high-resolution or volumetric OCT data. The reliance of many competitive architectures on hybrid CNN–transformer designs further suggests that pure attention models do not fully resolve local feature extraction limitations. Interpretability, while improved through attention visualization, remains incomplete. Attention maps do not always correspond to clinically relevant regions, and their reliability as explanations is still debated. As with CNN-based methods, the predominant reliance on 2D slices and the absence of large-scale prospective clinical validation remain systemic concerns shared across all supervised paradigms.

		\subsection{Hybrid CNN–Transformer Architectures}
		
		CNNs and pure transformers exhibit complementary weaknesses. CNNs are limited in modeling global context, while transformers struggle with fine-grained local feature extraction and require large amounts of labeled data. Hybrid architectures are motivated by the desire to address both limitations within a unified framework. The central premise is straightforward. CNNs excel at capturing low-level local patterns such as edges, textures, and small pathological structures, through their inductive spatial biases. Transformers provide coherent long-range dependency modeling through self-attention. Hybrid designs aim to combine both capabilities rather than sacrificing one for the other.
		In segmentation, works such as \cite{zhang2023transegnet,jiang2024hyformer} assign low-level feature extraction to CNN-based encoders while delegating global dependency modeling to transformer modules. This division of labor yields improved accuracy in delineating retinal layers and pathological regions, where local precision and structural consistency are jointly required. For disease classification, models including \cite{ma2022hctnet,yang2024hrs,laouarem2024htc} enrich feature representations by combining convolutional backbones with transformer blocks. These approaches generalize across imaging modalities, including fundus photography. Multi-scale and hierarchical feature integration, as seen in \cite{yang2024hrs,ashoka2025effivit}, enables simultaneous capture of micro-level lesions and macro-level structural patterns. This is particularly relevant given that retinal diseases often manifest across multiple spatial scales. Some works further incorporate attention-based visual explanations within hybrid ensemble frameworks, partially addressing the interpretability gap identified in both CNN and pure transformer approaches.
		Despite these advantages, hybrid architectures introduce new challenges. Combining convolutional layers, transformer blocks, and fusion mechanisms substantially increases model complexity. The resulting parameter counts are harder to train, tune, and deploy in resource-constrained clinical environments. There is also an absence of standardized design principles. Hybrid combinations are implemented sequentially, in parallel, or hierarchically depending on the work, complicating cross-method comparison and hindering identification of optimal strategies. Fine-grained localization can remain insufficient when attention mechanisms are not well aligned with clinically relevant structures, limiting segmentation reliability. The broader systemic limitations persist across all three supervised paradigms reviewed here: 2D-only operation, dataset-specific generalization, and the absence of prospective clinical validation. These are addressed more directly by the representation learning approaches discussed in subsequent sections.

		\section{Self-Supervised Representation Learning}
		\label{sec:ssl}
		The supervised paradigms described above share a fundamental dependence on large quantities of manually annotated data a requirement that is particularly burdensome in medical imaging, where expert labeling is time-consuming, expensive, and often inconsistent across clinicians. To overcome this annotation bottleneck, a growing body of work \cite{khosla2020supervised} explores Self-Supervised Learning (SSL), Semi-Supervised Learning, and domain adaptation strategies that leverage the abundance of unlabeled OCT data. A common thread across most proposed works \cite{rivail2019modeling} is the design of pretext tasks through which models learn meaningful representations without manual annotations, with the expectation that these representations transfer effectively to downstream clinical tasks.
		
		In the self-supervised setting, labels are unavailable or deliberately withheld during representation learning. The dataset is partitioned into a large unlabeled pool $\mathcal{D}_u = \{x_i\}_{i=1}^{N_u}$ and a small labeled set $\mathcal{D}_l = \{(x_i, y_i)\}_{i=1}^{N_l}$, with $N_l \ll N_u$. A pretext task $\mathcal{T}$ is constructed directly from the input data to generate surrogate supervision signals. Formally, for each sample $x_i$, the pretext task defines a pseudo-target $\tilde{y}_i = \mathcal{T}(x_i)$ (e.g., a predicted retinal thickness map, a pseudo-label from an alternative modality, or a transformed view of the same image, called also augmentation), and the encoder is trained by minimizing a self-supervised objective:
		
		\begin{equation}
			\min_{\theta} \; \mathbb{E}_{x \sim \mathcal{D}_u} \left[ \mathcal{L}_{\text{self}}\big(h_{\psi}(f_{\theta}(x)),\, \mathcal{T}(x)\big) \right]
		\end{equation}
		
		where $h_{\psi}$ is a lightweight projection head discarded after pretraining. The resulting encoder $f_\theta$ is subsequently fine-tuned on $\mathcal{D}_l$ for the downstream clinical task:
		
		\begin{equation}
			\min_{\theta, \phi} \; \mathbb{E}_{(x,y)\sim \mathcal{D}_l} \left[ \mathcal{L}\big(g_{\phi}(f_{\theta}(x)), y\big) \right]
		\end{equation}
		
		The quality of the learned representation $z = f_\theta(x)$ therefore depends critically on the alignment between the pretext task $\mathcal{T}$ and the semantic structure of the downstream clinical objective.

		Representative pretext tasks include the prediction of retinal thickness maps as a surrogate for anatomical structure learning \cite{holmberg2020self}, and iterative pseudo-label refinement schemes that progressively improve classification performance without ground truth annotations \cite{qiu2019self}. Temporal information is exploited in \cite{rivail2019modeling}, where consistency across longitudinal patient visits is used to model disease progression, marking a shift from static image analysis toward time-aware modeling relevant for chronic retinal conditions. Multimodal self-supervision is explored in \cite{hervella2019self,li2020self}, where cross-modal consistency between complementary imaging modalities generates pseudo-supervisory signals, enriching feature representations while reducing annotation dependence. Complementary directions include self-supervised denoising without clean reference images \cite{gisbert2020self}, clustering-based objectives for unsupervised retinal pattern discovery \cite{luo2020retinal}, cross-dataset domain adaptation without labeled target data \cite{he2020self}, and adversarial semi-supervised strategies combining labeled and unlabeled data to improve segmentation \cite{liu2018semi}.
		
		Despite their appeal, self-supervised methods introduce limitations that partially echo those of supervised approaches while raising new concerns specific to the unsupervised setting. Particularly contrastive approaches rely on the assumption that augmentations preserve semantic content, which can be mathematically problematic: if the transformation group or the pretext task $\mathcal{T}$ does not respect the true data manifold, the learned invariances may discard clinically relevant information. While this assumption is reasonable for natural images under transformations like cropping, flipping, or color jittering, it becomes problematic in OCT imaging. In fact certain transformations such as intensity scaling, spatial distortion, or aggressive cropping can alter or remove subtle pathological patterns (e.g., fluid regions or layer disruptions), thereby violating the semantic consistency assumption.
		
		Therefore pretext task design is critical: surrogate objectives misaligned with clinical goals may yield representations that miss pathologically relevant features. The annotation bottleneck is alleviated rather than eliminated, as fine-tuning on labeled data is still typically required, with performance often degrading under extreme label scarcity. Generalization across devices, populations, and acquisition protocols remains challenging \cite{wang2020domain,he2020self,chai2020perceptual}, and the predominant reliance on 2D processing neglects the volumetric structure of OCT data. This limitation motivates the 3D approaches discussed later. Error propagation in pseudo-labeling, training instability in adversarial frameworks, and methodological complexity including multi-stage pipelines and hyperparameter sensitivity further constrain reproducibility and clinical adoption. These challenges motivate a complementary family of approaches that address annotation scarcity by explicitly modeling the data distribution rather than designing pretext tasks, as reviewed in the following section.
		
		\section{Generative Representation Learning}
		\label{sec:generative}
		While self-supervised approaches reduce annotation dependency through pretext tasks, a complementary family of methods addresses the same challenge by explicitly modeling the underlying data distribution. Generative Representation Learning encompasses several paradigms from classical autoencoders to adversarial and diffusion-based models each offering distinct mechanisms to learn structured latent representations of retinal images in an unsupervised or semi-supervised manner.
		
		Generative approaches frame representation learning as a problem of modeling the underlying data distribution $p_{\text{data}}(x)$ of OCT images. Given that observations are corrupted by acquisition noise,  $x = x^\star + \epsilon$, where $x^\star$ denotes the latent clean anatomical structure, the generative objective is to learn a probabilistic model $p_\theta(x)$ that captures the true data distribution, from which clean reconstructions and structured latent representations can be derived.
		
		\subsection{Autoencoders and Variational Models}
		
		Autoencoders and Variational Autoencoders (VAEs) \cite{asano2020visualizing} constitute the foundational generative paradigm applied to retinal OCT analysis. The central principle is that latent variable models can encode compact and structured representations of retinal anatomy and pathology, enabling downstream tasks including clustering, classification, denoising, and segmentation without requiring fully supervised training.
		
		VAEs introduce a latent variable $z \in \mathcal{Z}$ and learn an approximate posterior $q_\phi(z \mid x)$ alongside a generative decoder $p_\theta(x \mid z)$, optimizing the Evidence Lower Bound (ELBO):
		
		\begin{equation}
			\max_{\theta, \phi} \; \mathbb{E}_{x \sim \mathcal{D}} \left[ \mathbb{E}_{q_\phi(z|x)}\big[\log p_\theta(x \mid z)\big] - D_{\mathrm{KL}}\big(q_\phi(z \mid x) \;\|\; p(z)\big) \right]
		\end{equation}
		
		where $p(z) = \mathcal{N}(0, I)$ is a standard Gaussian prior and $D_{\mathrm{KL}}$ denotes the Kullback–Leibler divergence. The encoder $f_\theta(x) = \mu_\phi(x)$ maps each OCT image to a structured latent code, which can be used for downstream clustering, classification, or segmentation.

		Several studies leverage VAEs to project high-dimensional OCT images into lower-dimensional latent spaces where disease characterization becomes tractable, with the expectation that latent embeddings capture clinically meaningful variation across pathological conditions. Other works \cite{laves2019retinal} integrate VAE frameworks into supervised classification pipelines as regularizers, which enforces feature space smoothness and improving generalization in low-data regimes.
		
		Autoencoder-based methods have been widely applied to denoising and image enhancement \cite{laves2019semantic,biswas2019dvae,asaoka2020usefulness}, exploiting reconstruction objectives to suppress speckle noise and acquisition artifacts without requiring paired clean targets. VAE-based approaches have further been extended to temporal modeling \cite{berchuck2019estimating}, using latent representations to predict disease progression and future clinical outcomes, and to semi-supervised segmentation \cite{sedai2017semi}, where unlabeled data is incorporated by constraining the latent space to improve performance under limited annotation.
		
		VAEs are conceptually appeal. They, however, exhibit well-documented limitations. Their probabilistic reconstruction objective tends to favor global consistency over high-frequency detail, producing overly smooth and blurry outputs \cite{doersch2016tutorial}. This is a significant drawback in medical imaging where fine structural details are diagnostically essential. Latent space interpretability is also limited: although VAEs enforce a structured representation, individual latent dimensions do not necessarily correspond to meaningful anatomical or pathological factors \cite{xue2022perioperative,sarhan2019learning,houssein2025explainable}. Posterior collapse \cite{he2019lagging}, whereby the decoder ignores latent variables, further undermines representation quality, particularly when expressive decoders are used. Training stability is sensitive to the balance between reconstruction loss and KL divergence regularization. As with supervised and self-supervised approaches, the predominant reliance on 2D representations limits exploitation of volumetric OCT structure \cite{pan2021retinal,kulyabin2024octdl}, and validation on small, homogeneous datasets constrains generalization. These limitations motivate the adoption of more expressive generative frameworks, as discussed below.
		
		\subsection{GAN-based Models}
		
		Generative Adversarial Networks (GANs) \cite{goodfellow2020generative} extend the generative paradigm by replacing VAEs' probabilistic reconstruction objective with an adversarial training scheme, driving the generator to produce sharper and more realistic outputs directly addressing the blurriness that characterizes VAE-based synthesis. Applied to OCT imaging, GAN-based approaches target data scarcity, noise, and rare pathology detection through the core principle that learning to generate realistic retinal data provides a better understanding of both normal and pathological image distributions.
		
		GANs replace the reconstruction objective with an adversarial game between a generator $G_\theta : \mathcal{Z} \rightarrow \mathcal{X}$ and a discriminator $D_\phi : \mathcal{X} \rightarrow [0,1]$, optimized through the minimax objective:
		
		\begin{equation}
			\min_{\theta} \max_{\phi} \; \mathbb{E}_{x \sim p_{\text{data}}} \big[\log D_\phi(x)\big] + \mathbb{E}_{z \sim p(z)} \big[\log(1 - D_\phi(G_\theta(z)))\big]
		\end{equation}
		
		This formulation encourages $G_\theta$ to produce sharp, perceptually realistic OCT images whose distribution is indistinguishable from $p_{\text{data}}(x)$, addressing the blurriness inherent to VAE reconstruction.
		
		This principle is exploited across several application directions. In anomaly detection, \cite{zhou2020sparse} identifies pathological regions through reconstruction errors on a model trained exclusively on normal OCT images, while \cite{zhang2020memory} strengthens this approach with memory modules that improve sensitivity to subtle deviations. Image denoising and super-resolution are addressed in \cite{chen2020dn,kande2020siamesegan,hasan2021deep} and \cite{das2020unsupervised} respectively, where adversarial training yields perceptually sharper results than traditional filtering approaches. For data augmentation, \cite{xiao2020open} synthesizes samples to improve robustness to unseen classes, \cite{he2020retinal} uses generated data as a classifier regularizer, and \cite{sun2020gan} performs cross-dataset image translation to mitigate domain shift. Semi-supervised segmentation frameworks such as \cite{lee2020retinal,park2020m} employ discriminators to enforce consistency between predicted and real segmentation maps, leveraging unlabeled data within the adversarial paradigm. The breadth of these applications is surveyed in \cite{sengupta2020deep}, which highlights GANs as a versatile unifying framework across classification, segmentation, and enhancement.
		
		GANs, however, introduce limitations that compound those of VAEs while adding new ones. Training instability, mode collapse, and hyperparameter sensitivity are well-documented challenges \cite{kammoun2022generative,karras2020training} that affect reproducibility. Visual realism does not guarantee clinical validity: generated images may contain hallucinated or anatomically incorrect structures, a concern shared with VAE-based synthesis but potentially more consequential given GANs' higher output fidelity. Evaluation via generic metrics such as PSNR and SSIM \cite{salehi2020generative,treder2022quality} does not reliably reflect diagnostic utility. Anomaly detection frameworks \cite{schlegl2019f,zhou2020sparse} are sensitive to underrepresentation of benign variation in the normal training distribution, generating false positives. Synthetic augmentation may introduce distributional biases, and domain adaptation via GANs can fail to preserve clinically relevant features when source and target domains differ substantially. The limitations of 2D processing, limited interpretability, and insufficient large-scale clinical validation persist, motivating the shift toward diffusion-based generative models.
		
		\subsection{Diffusion-Based Models}
		Diffusion models \cite{croitoru2023diffusion,cao2024survey} represent the most recent advance in generative representation learning for retinal OCT analysis, addressing key shortcomings of both VAEs and GANs  most notably training instability and output quality. Rather than relying on adversarial objectives or variational encoders, diffusion models learn the data distribution through a progressive denoising process: Gaussian noise is iteratively added to training images, and the model is trained to reverse this corruption, enabling high-fidelity synthesis from pure noise. This probabilistic formulation yields diverse, structurally consistent retinal images with greater training stability than GANs.
		
		Formally, diffusion models define a forward noising process $q(x_t \mid x_{t-1}) = \mathcal{N}(x_t;\, \sqrt{1-\beta_t}\, x_{t-1},\, \beta_t I)$ that progressively corrupts a clean image $x_0 \sim p_{\text{data}}$ over $T$ steps, and learn a parameterized reverse process $p_\theta(x_{t-1} \mid x_t)$ to denoise from pure Gaussian noise:
		
		\begin{equation}
			\min_{\theta} \; \mathbb{E}_{x_0,\, \epsilon,\, t} \left[ \left\| \epsilon - \epsilon_\theta\!\left(x_t,\, t\right) \right\|^2 \right]
		\end{equation}
		
		where $\epsilon \sim \mathcal{N}(0, I)$ is the noise added at step $t$ and $\epsilon_\theta$ is the learned denoising network. High-fidelity OCT images are synthesized by iteratively applying the reverse process from $x_T \sim \mathcal{N}(0, I)$, yielding superior sample quality and training stability compared to both VAEs and GANs.

		Applications span several clinically relevant directions. Synthetic data augmentation for segmentation, demonstrated in \cite{garcia2024using,li2025retidiff}, enriches training distributions and directly addresses annotation scarcity, while \cite{wu2024retinal,yamauchi2025evaluation} show improved robustness under limited annotation regimes. Cross-modality translation, inferring OCTA vascular information from structural OCT scans \cite{badhon2025diffusion,rashid2024using}, reduces the clinical burden of multi-modal acquisition. Image restoration applications include super-resolution for portable devices \cite{tian2025octdiff}, artifact removal \cite{ji2023novel}, and general denoising \cite{ersari2025denoising}. Controllable synthesis guided by anatomical or pathological conditions \cite{de2025controllable,liu2026tortuosity} and text-conditioned generation for rare disease cases \cite{chen2024eyediff} extend the framework toward interactive clinical tools.
		
		Diffusion models, however, inherit several limitations from earlier generative paradigms while introducing new ones. Computational cost is substantially higher than GANs or VAEs \cite{fuest2026diffusion}: the iterative denoising process requires many forward passes, limiting real-time clinical applicability. Hallucinated or anatomically incorrect structures remain a risk, particularly under data-limited training conditions. Evaluation methodology shares the weaknesses observed across generative approaches, with visual metrics poorly reflecting clinical utility. Cross-modality outputs such as OCT-to-OCTA synthesis risk misrepresenting physiological signals, potentially misleading vascular interpretation. Controllability and interpretability remain limited despite conditioning mechanisms \cite{de2025controllable}, the 2D processing limitation persists, and most methods lack large-scale multi-center clinical validation. Overall, diffusion models represent a meaningful advancement over earlier generative paradigms in terms of output quality and training stability, but computational demands, clinical reliability, and scalability must be resolved for practical deployment. These unresolved challenges, in particular the persistent neglect of OCT's volumetric nature, motivate the paradigm shift reviewed in the following section.
		
		\section{3D and Volumetric Representation Learning}
		\label{sec:3d}
		A limitation consistently identified across all paradigms reviewed so far supervised, self-supervised, and generative is the predominant reliance on 2D slice-based processing, which discards the inter-slice correlations inherent to volumetric OCT data. Addressing this gap, a growing body of work advocates for a shift toward 3D volumetric analysis, grounded in the observation that OCT volumes form a continuous 3D representation of retinal structures whose spatial coherence carries diagnostically critical information not recoverable from independently processed B-scans.

		Volumetric representation learning extends the general framework by replacing the 2D input $x_i \in \mathbb{R}^{H \times W}$ with a full 3D OCT volume $x_i \in \mathbb{R}^{H \times W \times D}$, where $D$ denotes the number of B-scans in the acquisition stack. The dataset is usually defined as $\mathcal{D} = \{(x_i, y_i)\}_{i=1}^N$ with $x_i \in \mathbb{R}^{H \times W \times D}$ and $y_i$ denoting a volumetric annotation (e.g., a 3D segmentation mask or a scalar clinical variable). The encoder $f_\theta : \mathbb{R}^{H \times W \times D} \rightarrow \mathcal{Z}$ must capture not only intra-slice spatial features but also inter-slice dependencies encoding structural continuity across the retinal volume. The learning objective retains the general form in Equation~\ref{eq:representationLearningSupervised}.
		% \begin{equation}
			% \min_{\theta, \phi} \; \mathbb{E}_{(x,y)\sim \mathcal{D}} \left[ \mathcal{L}\big(g_{\phi}(f_{\theta}(x)), y\big) \right]
			% \end{equation}
		
		However, the anisotropic resolution of OCT volumes where in-plane pixel spacing $(\Delta h, \Delta w)$ typically differs from inter-slice spacing $\Delta d$ must be explicitly accounted for in the design of $f_\theta$. Convolutional kernels or attention mechanisms applied uniformly across all three dimensions without correction for this anisotropy may produce geometrically inconsistent representations. Formally, the noise model is extended to:
		
		\begin{equation}
			x = x^\star + \epsilon^{3D}
		\end{equation}
		
		where $\epsilon^{3D}$ accounts for both intra-slice speckle noise and inter-slice motion artifacts that propagate spatially across the volume, making the recovery of $x^\star$ a volumetric denoising and reconstruction problem.
		
		Key directions in this space include 3D reconstruction and visualization \cite{aaker2011volumetric,lopez2022fully}, generating coherent volumetric representations that support spatial interpretation of disease morphology beyond what 2D renderings allow. 
		
		Quantitative vascular analysis in three dimensions, explored in \cite{zhang20193d,feu2024retinal}, provides biomarkers such as vessel density and volume that are inherently richer than 2D projections. Microvasculature modeling using OCTA \cite{sarabi20203d,liu2024ai} captures blood flow and vascular network topology volumetrically, yielding more accurate and clinically relevant measurements for conditions such as diabetic retinopathy and macular degeneration. Volumetric segmentation methods \cite{mukherjee2022retinal,sleman2021novel} extend convolutional architectures to 3D to delineate retinal layers and pathological regions while preserving spatial continuity, and \cite{cahyo2021multi} further combines segmentation with structural reconstruction objectives. Longitudinal analysis, as in \cite{borrelli2022longitudinal}, demonstrates that volumetric data across time enables integrated spatio-temporal modeling of disease progression. This direction is only partially addressed by the temporal self-supervised approaches reviewed earlier.
		
		The challenges introduced by 3D modeling are substantial and largely orthogonal to those of 2D paradigms. Computational cost increases dramatically: volumetric processing demands far greater memory and processing power, constraining scalability to high-resolution acquisitions and limiting real-time applicability. The annotation bottleneck is severely exacerbated, as manual labeling of 3D volumes is considerably more complex and time-consuming than 2D annotation, restricting training dataset sizes. Architectural design must contend with anisotropic resolution (the typically different in-plane and inter-slice spacings in OCT). Neglecting this factor can substantially degrade model performance. Noise and artifacts including speckle and motion, already challenging in 2D, propagate across slices in volumetric settings and compound their impact on analysis quality. Clinical usability of 3D outputs is also non-trivial, as volumetric renderings can be difficult for clinicians accustomed to 2D slice review to interpret effectively. Standardized evaluation protocols for 3D methods remain lacking, with heterogeneous datasets, metrics, and preprocessing pipelines impeding cross-method comparison. In summary, volumetric modeling represents a necessary step forward for retinal OCT analysis, but computational, annotation, and methodological challenges must be resolved before these approaches can be reliably translated to clinical practice.
		
		\section{Multimodal Representation Learning}
		\label{sec:multimodal}
		While 3D approaches address the spatial limitations of 2D slice-based processing, they remain largely restricted to a single imaging modality. A complementary line of work extends the representational scope further by integrating multiple heterogeneous data sources such as OCT, fundus imaging, and patient-specific clinical metadata. The motivating principle is that clinical decision-making inherently relies on combining diverse information streams, and that unimodal models therefore capture only a partial view of the diagnostic process regardless of their architectural sophistication.
		
		The benefit of metadata-enhanced learning is demonstrated in works such as \cite{holland2024metadata}, where auxiliary patient information including age, intraocular pressure, and other clinical variables, guides feature extraction and enriches the interpretability of imaging biomarkers. Li et al. \cite{li2023transfer} integrates OCT, fundus images, and demographic metadata within a unified model, achieving improved classification accuracy over unimodal baselines. Fusion architecture design is a central methodological concern: works including \cite{zedadra2025multi,liu2025oct} combine imaging and clinical data through early, late, or intermediate feature fusion strategies. They often leverage Vision Transformers for cross-modal interaction to predict disease presence or postoperative outcomes. The structural complementarity of OCT and fundus imaging is a recurring theme: OCT provides cross-sectional anatomical detail while fundus imaging offers a global retinal view, and their joint exploitation enables more comprehensive disease characterization. Automated report generation, where imaging features are combined with clinical context to produce structured textual summaries, represents a natural extension of multimodal integration toward end-to-end clinical communication systems. \cite{arunga2024assessment} additionally highlights the role of multimodal data in improving annotation consistency during dataset curation, suggesting utility beyond model training. Foundation models such as \cite{morano2025multimodal} aim to learn unified representations transferable across classification, segmentation, and report generation. This convergence between multimodal learning and the large-scale pretraining paradigm is discussed in the following section.
		
		Multimodal learning, however, introduces challenges that compound those identified in prior sections. Data synchronization is a fundamental practical obstacle, as missing modalities, inconsistent acquisition protocols, and incomplete metadata records are endemic to clinical datasets and can substantially degrade model performance. Fusion strategy design remains empirically driven, with no established consensus on optimal combination approaches. This form of architectural heterogeneity parallels the lack of standardization observed in hybrid CNN–Transformer models. Clinical metadata introduces variability in format, quality, and completeness that can inject noise and bias into learned representations. The annotation bottleneck is further aggravated, as assembling and curating multimodal datasets is considerably more demanding than single-modality equivalents. Privacy regulations and data governance constraints restrict cross-institutional sharing, limiting dataset diversity and scale. Disentangling individual modality contributions to final predictions remains difficult, echoing the interpretability challenges recurring throughout this review. Generalization across clinical settings with varying devices, protocols, and populations, and the persistent absence of prospective multi-center validation, represent shared concerns with all prior paradigms. These limitations underscore that multimodal integration, while clinically compelling, does not resolve the fundamental challenges of data availability and evaluation rigor.
		
		\section{Foundation Models and Pretraining}
		\label{sec:foundation}
		The trajectory across the preceding sections from task-specific supervised models, through self-supervised and generative approaches, to multimodal integration converges toward a broader ambition: the development of general-purpose foundation models capable of addressing multiple tasks, modalities, and interaction modes within a single large-scale pretrained framework. Rather than engineering solutions for individual problems, these approaches leverage massive pretraining corpora to learn universal representations of ophthalmic data that can be efficiently adapted to diverse downstream tasks, representing a qualitative shift in how AI systems for retinal analysis are conceived and built.
		
		A central mechanism is the alignment of visual and linguistic representations through contrastive learning, enabling cross-modal understanding between retinal images and textual clinical descriptions. Models such as \cite{shi2024eyeclip,shi2024eyefound,morano2025multimodal} extend this paradigm to ophthalmic data, learning joint embeddings that support classification, retrieval, and zero-shot generalization across tasks not seen during pretraining. Language integration further enables clinical reasoning capabilities: systems such as \cite{jalili2025glaucoma,da2025integrated} generate structured diagnostic reports, respond to clinical queries, and support patient triage. This moves beyond passive image analysis toward interactive AI systems embedded in real clinical workflows. Benchmarking infrastructure developed in works such as \cite{wang2024multieye,liang2025novel} provides standardized evaluation protocols across modalities and tasks. This establishes the reproducibility foundations necessary for progress tracking in this rapidly evolving field. Domain-specific adaptation strategies highlighted in \cite{holland2025specialized} address the recognized gap between generic pretraining and the fine-grained structural specificity of retinal pathology.
		
		Foundation models, however, concentrate and amplify many of the limitations identified across all prior paradigms. Data requirements reach a new order of magnitude: effective large-scale pretraining demands diverse, well-curated datasets that are exceptionally difficult to assemble in medical imaging given privacy constraints, regulatory barriers, and the annotation bottleneck that motivated earlier sections. Computational cost similarly escalates beyond that of transformer-based or diffusion-based models, raising serious accessibility concerns for resource-limited clinical environments. The domain specificity gap is critical: generic visual representations may fail to encode the diagnostically decisive fine-grained pathological details (subtle fluid accumulation, microstructural disruptions, vascular anomalies) that distinguish retinal conditions, and targeted fine-tuning reintroduces labeled data dependence. The interpretability problem, recurring across every paradigm in this review, reaches its most acute form in large vision-language models, where generated explanations are not always clinically faithful, reasoning processes remain largely opaque, and hallucination poses direct risks in diagnostic contexts. Evaluation frameworks adequate for multi-task, multi-modal foundation models remain underdeveloped, with existing metrics insufficiently capturing clinical safety or utility. Bias and fairness concerns are amplified at scale, as models trained on demographically skewed data may systematically underperform on underrepresented populations. The generality–specialization trade-off further complicates deployment decisions: models optimized for broad task coverage may not match dedicated architectures on specific clinical benchmarks. In summary, foundation models represent the current frontier of retinal OCT analysis, synthesizing the strengths of supervised learning, self-supervised pretraining, generative modeling, multimodal fusion, and language-grounded reasoning within a unified framework. Yet their safe, equitable, and clinically reliable deployment depends on resolving challenges in data, interpretability, evaluation, and domain adaptation that remain fundamentally open across the field as a whole.
		
		\section{Challenges and Open Research Directions}
		\label{sec:challenges}
		Despite substantial progress across the paradigms reviewed in this survey, a comprehensive analysis reveals a set of persistent and interrelated challenges that collectively hinder the deployment of robust and clinically reliable systems. These challenges are systemic. They recur across architectures, learning paradigms, and application domains. At the same time, they directly point toward the most important open research directions for the field.
		
		\subsection{Data Limitations and Dataset Diversity}
		
		The scarcity and limited diversity of annotated OCT datasets constitute perhaps the most pervasive bottleneck in the field. OCT annotation requires expert ophthalmologists. This makes labeling costly, time-consuming, and difficult to scale. Even widely used datasets typically cover only a small number of disease categories and a restricted patient population. This scarcity directly causes overfitting in data-hungry architectures such as Vision Transformers, GANs, and diffusion models. It also constrains the diversity of pathological patterns that models are exposed to during training. Although self-supervised, semi-supervised, and generative approaches aim to alleviate this bottleneck, they reduce rather than eliminate annotation dependence. Labeled data remains necessary for fine-tuning and clinical evaluation.
		
		Compounding scarcity is the problem of dataset bias and lack of representativeness. The majority of OCT datasets are collected from a single institution, acquired with a single imaging device, and represent limited demographic variability. Models trained under these conditions tend to perform well on benchmark splits while failing to generalize to external datasets, rare disease presentations, or underrepresented patient populations. Evaluation methodology further contributes to inflated performance estimates. The widespread use of random rather than patient-wise data splits allows B-scan-level leakage across training and test sets, producing overly optimistic results that do not reflect real-world generalization.
		
		A third frequently overlooked data challenge is annotation noise and inconsistency. Inter-observer variability among clinicians, ambiguous borderline cases, and incomplete clinical context at labeling time introduce systematic label noise. This noise propagates through both supervised and self-supervised pipelines. In pseudo-labeling and iterative refinement schemes, early annotation errors can compound through training, undermining model reliability in ways that are difficult to detect from benchmark metrics alone.
		
		Addressing these data limitations requires a coordinated effort. Federated and privacy-preserving representation learning frameworks~\cite{kaissis2020secure,koutsoubis2024privacy,koutsoubis2025privacy} offer a principled path toward leveraging data across institutions without sharing sensitive patient records. Federated learning~\cite{lo2021federated,nguyen2022federated,nabil2025federated} has been demonstrated for OCT-based disease classification and OCTA microvasculature segmentation across multiple sites. However, combining federated training with self-supervised or foundation model pretraining remains largely unexplored. This combination constitutes a direct path toward the large-scale, demographically diverse training corpora that foundation models require. Closely related is the challenge of algorithmic fairness and bias mitigation~\cite{nakayama2023fairness,tian2024fairdomain,xu2024addressing}. Models trained predominantly on data from specific geographic regions, demographic groups, or OCT device manufacturers may systematically underperform on underrepresented populations. The literature offers very limited evidence that current OCT models have been evaluated for equity across demographic subgroups. Establishing fairness-aware training objectives and evaluation protocols as standard practice is an urgent open problem with direct implications for equitable clinical deployment.
		
		\subsection{Representation Learning Deficiencies}
		
		A fundamental limitation shared across all reviewed paradigms is the opacity of learned representations and their disconnect from clinical reasoning. Despite advances in visualization techniques including saliency maps and Grad-CAM in CNN-based approaches~\cite{ayhan2024interpretable,thakoor2020robust} and attention map visualization in transformer-based models~\cite{he2023interpretable,playout2022focused}, most models remain effectively black boxes from a clinical perspective. Post-hoc explanations are frequently coarse, low-resolution, and spatially imprecise. They highlight broad image regions rather than the specific pathological structures that clinicians rely on diagnostically. These include fluid accumulation subtypes, specific layer disruptions, and vascular morphological features. Critically, post-hoc explanations are applied after training and do not reflect the model's internal reasoning process. Attention mechanisms in transformers do not reliably indicate feature importance, and their clinical meaningfulness remains contested. Approaches that incorporate anatomical and physiological domain knowledge~\cite{lux2025interpretable} partially address this gap but typically at the cost of increased complexity and reduced scalability. In multimodal settings~\cite{mehta2021automated}, disentangling the contributions of individual modalities to a prediction presents an additional layer of complexity. Generative models including VAEs, GANs, and diffusion models further fail to provide interpretable latent spaces. Their latent variables are abstract statistical constructs not directly linked to clinical concepts. The absence of standardized quantitative metrics for interpretability evaluation means that most studies rely on qualitative visual assessments, making objective comparison across methods practically infeasible.
		
		The emerging direction of concept-based interpretability~\cite{doumanoglou2023unsupervised,wen2024concept} organizes representations explicitly around clinically meaningful semantic concepts. It offers a more principled alternative and is beginning to appear in the OCT literature, but remains far from mature. Closely related is the broader problem of weak clinical alignment of learned representations~\cite{sucholutsky2023getting}. Models tend to capture statistical regularities in pixel distributions rather than physiologically meaningful structures such as retinal layer boundaries or fluid biomarkers. This misalignment limits the utility of AI outputs in real clinical workflows and contributes to the persistent trust gap between clinicians and automated systems.
		
		Most current models also produce deterministic point predictions without reliable uncertainty estimates. This is a significant limitation in a medical context where confidence assessment is a prerequisite for safe clinical use. Approaches such as Bayesian deep learning and Monte Carlo dropout~\cite{dechesne2021bayesian} exist but are rarely integrated into OCT analysis pipelines. They typically incur high computational cost with limited scalability. Models may therefore produce overconfident predictions on ambiguous or out-of-distribution cases, precisely the situations where calibrated uncertainty would be most valuable. This limitation is particularly acute for generative anomaly detection approaches, where the absence of uncertainty quantification makes it difficult to distinguish genuine pathological findings from distribution artifacts. A recent work~\cite{peng2025enhancing} introduced a foundation model with explicit uncertainty estimation capable of detecting 16 retinal conditions on OCT. It demonstrated that uncertainty-aware models not only achieve higher diagnostic accuracy but also flag ambiguous or low-quality inputs for manual review. Integrating uncertainty estimation directly into the representation learning objective rather than as a post-hoc addition remains an open methodological challenge.
		
		\subsection{Methodological Gaps and Open Directions}
		
		The predominant reliance on 2D slice-based processing represents a structural methodological gap across nearly all reviewed paradigms. OCT data is inherently volumetric. Inter-slice correlations encode structural continuity and disease extent information that cannot be recovered from independently processed B-scans. While 3D CNN and volumetric approaches~\cite{russakoff20203d,gessert2018force} exist, they remain computationally demanding and dataset-constrained. Critically, the intersection of volumetric modeling with self-supervised, generative, or foundation model pretraining is largely unexplored. Current ophthalmic foundation models such as RETFound~\cite{zhou2023foundation,chuter2025multimodal} and MIRAGE~\cite{morano2025multimodal} process only the central B-scan of an OCT volume. A recent study~\cite{judkiewicz2026shifting} demonstrated that adapting video foundation models such as V-JEPA \cite{bardes2023v,assran2025v} to treat OCT volumes as temporal sequences of B-scans achieves statistically significant improvements in AMD and glaucoma detection. This establishes video-based pretraining as a compelling pathway toward volumetric OCT understanding. Extending representation learning paradigms to 3D settings while accounting for anisotropic resolution, inter-slice motion artifacts, and the amplified annotation burden of volumetric labeling represents one of the most significant open challenges in the field.
		
		Longitudinal and spatiotemporal representation learning for disease progression modeling remains equally underdeveloped. Chronic retinal diseases such as AMD and diabetic retinopathy evolve over months and years. Representations that capture patient-level temporal trajectories across multiple OCT visits are essential for prognosis, treatment response prediction, and clinical trial outcome modeling. Yet the vast majority of reviewed approaches treat each OCT acquisition independently. Developing self-supervised objectives that leverage longitudinal consistency across visits is a frontier only beginning to be explored. Combining these with volumetric and multimodal information represents a natural and clinically motivated next step.
		
		The literature is also substantially fragmented across supervised, self-supervised, generative, transformer-based, multimodal, and foundation model paradigms. Few works combine self-supervised pretraining with 3D volumetric modeling. Generative models are rarely integrated with clinical reasoning pipelines. Multimodal fusion is largely decoupled from volumetric representations. This fragmentation results in redundant parallel efforts and missed opportunities for synergistic combination.
		
		The convergence of vision-language pretraining with ophthalmic clinical reasoning opens a further important research direction. Foundation models that jointly encode retinal images and free-text clinical reports can in principle support zero-shot generalization to new conditions, automated structured report generation, and interactive clinical question answering. Recent models such as FLAIR \cite{rojas2022clinical} and EyeFound \cite{shi2024eyefound} begin to explore these capabilities. However, they require substantially larger and more linguistically diverse paired image-text corpora than are currently available. The faithful alignment of generated clinical text with true image-level findings is a safety-critical open problem. Hallucinated but plausible-sounding descriptions pose direct risks in diagnostic contexts.
		
		Finally, a pervasive gap between methodological development and clinical deployment persists across the entire reviewed literature. The vast majority of studies optimize and report standard benchmark metrics such as accuracy, AUC, and Dice score. They do so without prospective clinical validation, multi-center evaluation, or integration into real hospital workflows. Human-AI interaction, real-time decision support requirements, and regulatory compliance considerations are almost entirely absent. Evaluation protocols are further weakened by the inconsistencies noted above. Patient-wise splits, external validation cohorts, and standardized preprocessing pipelines are not uniformly adopted. The absence of universally accepted benchmarks for OCT representation learning represents a foundational infrastructure gap.
		Beyond these structural issues, a deeper and less discussed problem concerns the nature of the representations themselves. Current representation learning frameworks are designed to optimize statistical objectives on labeled datasets. They are not designed to replicate the cognitive process through which a clinician learns to interpret OCT images. A clinician does not learn from millions of labeled examples. They learn progressively, under supervision, by integrating visual pattern recognition with anatomical knowledge, physiological understanding, and years of diagnostic experience. This learning process is deeply contextual. It is guided by expert feedback, clinical outcomes, and the subtle interplay between image features and patient history. The representations learned by current models, however powerful statistically, do not encode this kind of structured clinical expertise. They capture correlations in pixel distributions rather than the diagnostically meaningful abstractions that experienced ophthalmologists rely on. This suggests a fundamental limitation that benchmark performance alone cannot reveal. A model may achieve high accuracy on a held-out test set while relying on spurious features that no clinician would recognize as clinically relevant.
		Addressing this limitation may require a fundamental rethinking of how representations are learned. One promising direction is to design learning objectives and training curricula that more closely mirror the way clinicians acquire expertise. This could involve learning from structured clinical feedback rather than fixed labels. It could involve incorporating anatomical and physiological priors directly into the representation learning objective. It could also involve developing interactive learning frameworks where clinician knowledge actively shapes the feature space during training rather than being applied only at evaluation time. Concept-based representation learning, where the latent space is explicitly organized around clinically meaningful semantic units, is one step in this direction. But much remains to be done. The most effective representations for clinical OCT analysis may ultimately be those that are not merely learned from clinical data but learned in the same way that clinicians learn: progressively, interactively, and under the guidance of structured domain expertise.
		Closing the translation gap therefore requires more than collaboration between the machine learning and clinical ophthalmology communities, although that collaboration is essential. It requires a reconceptualization of what a clinically useful representation should be. It requires regulatory engagement and the establishment of shared evaluation infrastructure. It requires moving beyond accuracy metrics toward assessments of clinical utility, safety, and trustworthiness. This is arguably among the most consequential open direction of all.
		
		\section{Conclusions}
		\label{sec:conclusion}
		This survey has provided a comprehensive review of representation learning methods for retinal OCT image analysis. We have covered the full spectrum of approaches. This includes supervised CNN-based and transformer-based architectures, self-supervised and semi-supervised methods, generative models, 3D volumetric approaches, multimodal frameworks, and large-scale foundation models. Each paradigm has addressed specific limitations of its predecessors. Each has also introduced new challenges of its own. Together, they trace a clear trajectory. Learned representations have evolved from discriminative local features in individual B-scans toward universal multimodal embeddings supporting generalization across tasks, modalities, and clinical settings.
		Several key conclusions emerge from this analysis.
		The annotation bottleneck remains the most fundamental constraint in the field. Despite significant methodological progress, dependence on expert-labeled data has been reduced but not eliminated. Self-supervised, semi-supervised, and generative approaches have made meaningful steps toward leveraging unlabeled data. Foundation model pretraining represents the most promising current direction toward annotation-efficient learning. Yet labeled data remains necessary for fine-tuning and clinical evaluation. The cost of volumetric annotation continues to limit dataset diversity and scale.
		The reliance on 2D B-scan processing represents a persistent structural gap. OCT is inherently a volumetric modality. Inter-slice correlations carry diagnostically critical information. No 2D model can fully exploit this information. Transitioning to scalable 3D representation learning remains one of the most important open technical challenges in the field.
		Interpretability has not kept pace with predictive performance. State-of-the-art models achieve expert-level accuracy on benchmark tasks. Yet their decision-making remains largely opaque. Post-hoc visualization tools produce explanations that are coarse and often clinically uninformative. Representations intrinsically aligned with anatomical structures and pathological biomarkers are needed. This alignment is a prerequisite for clinical trust and regulatory acceptance.
		The evaluation infrastructure of the field requires urgent improvement. Universally accepted benchmarks do not yet exist. Random data splits remain widespread. Evaluation on single-center and single-device data dominates the literature. These practices inflate reported performance and obscure true generalization ability. Standardized, clinically meaningful evaluation protocols are as important as architectural advances.
		The convergence of multimodal learning, large-scale pretraining, and language-grounded reasoning marks the most significant paradigm shift currently underway. Foundation models jointly trained on retinal images and clinical text offer the prospect of unified systems. These systems could support classification, segmentation, report generation, and clinical decision support within a single framework. Realizing this potential requires resolving challenges in data diversity, domain adaptation, uncertainty quantification, and regulatory validation. The current literature has only begun to address these issues.
		Looking forward, the most impactful advances will likely emerge from the intersection of paradigms. Volumetric self-supervised pretraining combined with foundation models is one such intersection. Federated learning enabling cross-institutional collaboration without privacy compromise is another. Uncertainty-aware representations and longitudinal modeling of disease progression are equally important.

\section*{Acknowledgement}
This work has been carried out in the OCTIPA project (CMCU
23G1418), as part of the PHC-Utique program managed by
the CMCU of the French Ministry of Europe and Foreign
Affairs and the Tunisian Ministry of Higher Education
and Scientific Research



%Bibliography
\bibliographystyle{unsrt}  
\bibliography{referencebib}

\end{document}